\documentclass{article}
\usepackage[final]{nips_2017}

\usepackage{hyperref}
\usepackage{url}
\usepackage{multibib}
\usepackage{subfigure}
\usepackage{amsmath}
\usepackage{amsthm}
\usepackage{amsfonts}
\usepackage{amssymb}
\usepackage{graphicx}
\usepackage{color}
\usepackage{xcolor}
\usepackage{graphicx}
\usepackage{multirow}

\usepackage{algorithmic}
\usepackage{algorithm}
\newcommand{\dkls}[3]{\mathbb{D}_{KL}^{#1}[#2 \, \|\, #3]}

\newcommand{\sg}[2]{\widehat{\nabla}_{#2} #1}

\newcommand\cut[1]{}

\usepackage{algorithm}

\newcommand{\squishlist}{
   \begin{list}{$\bullet$}
    { \setlength{\itemsep}{0pt}      \setlength{\parsep}{3pt}
      \setlength{\topsep}{3pt}       \setlength{\partopsep}{0pt}
      \setlength{\leftmargin}{1.5em} \setlength{\labelwidth}{1em}
      \setlength{\labelsep}{0.5em} } }

\newcommand{\squishlisttwo}{
   \begin{list}{$\bullet$}
    { \setlength{\itemsep}{0pt}    \setlength{\parsep}{0pt}
      \setlength{\topsep}{0pt}     \setlength{\partopsep}{0pt}
      \setlength{\leftmargin}{2em} \setlength{\labelwidth}{1.5em}
      \setlength{\labelsep}{0.5em} } }

\newcommand{\squishend}{
    \end{list}  }
 
{}
{}
{}
{}
{}

\newcommand{\half}{\mbox{$\frac{1}{2}$}}

\newcommand{\rnd}[1]{\left(#1\right)}
\newcommand{\sqr}[1]{\left[#1\right]}
\newcommand{\crl}[1]{\left\{#1\right\}}
\newcommand{\myexpect}{\mathbb{E}}

\newcommand{\gauss}{\mbox{${\cal N}$}}

\newcommand{\myvec}[1]{\mbox{$\mathbf{#1}$}}
\newcommand{\myvecsym}[1]{\mbox{$\boldsymbol{#1}$}}

\newcommand{\vone}{\mbox{$\myvecsym{1}$}}

\newcommand{\vepsilon}{\mbox{$\myvecsym{\epsilon}$}}

\newcommand{\vmu}{\mbox{$\myvecsym{\mu}$}}
\newcommand{\vlambda}{\mbox{$\myvecsym{\lambda}$}}

\newcommand{\vphi}{\mbox{$\myvecsym{\phi}$}}

\newcommand{\vtheta}{\mbox{$\myvecsym{\theta}$}}

\newcommand{\vsigma}{\mbox{$\myvecsym{\sigma}$}}
\newcommand{\vSigma}{\mbox{$\myvecsym{\Sigma}$}}

\newcommand{\vg}{\mbox{$\myvec{g}$}}

\newcommand{\vm}{\mbox{$\myvec{m}$}}

\newcommand{\vs}{\mbox{$\myvec{s}$}}

\newcommand{\vx}{\mbox{$\myvec{x}$}}

\newcommand{\vy}{\mbox{$\myvec{y}$}}

\newcommand{\vA}{\mbox{$\myvec{A}$}}

\newcommand{\vI}{\mbox{$\myvec{I}$}}

\newcommand{\vS}{\mbox{$\myvec{S}$}}

\newcommand{\vX}{\mbox{$\myvec{X}$}}
\newcommand{\diag}{\mbox{$\mbox{diag}$}}

\newcommand{\ee}{\end{equation}}
\newcommand{\bea}{\begin{eqnarray}}
\newcommand{\eea}{\end{eqnarray}}
\newcommand{\beaa}{\begin{eqnarray*}}
\newcommand{\eeaa}{\end{eqnarray*}}

\newcommand{\argmin}{\mathop{\mathrm{argmin}}}
\newcommand{\argmax}{\mathop{\mathrm{argmax}}}

\usepackage[draft]{todonotes}   % notes showed
\title{Vprop: Variational Inference using RMSprop}

\author{Mohammad Emtiyaz Khan\\
RIKEN, Tokyo, Japan\\
\texttt{emtiyaz.khan@riken.jp}\\
\And 
Zuozhu Liu\thanks{Work done during an internship in RIKEN.}\\
SUTD, Singapore\\
\texttt{zuozhu\_liu@mymail.sutd.edu.sg}\\
\And
Voot Tangkaratt\\
RIKEN, Tokyo, Japan\\
\texttt{voot.tangkaratt@riken.jp}\\
\And 
Yarin Gal\\
University of Oxford, UK\\
\texttt{yarin.gal@cs.ox.ac.uk}
}

\begin{document}

\maketitle

\begin{abstract}
   Many computationally-efficient methods for Bayesian deep learning rely on continuous optimization algorithms, but the implementation of these methods requires significant changes to existing code-bases. In this paper, we propose Vprop, a method for Gaussian variational inference that can be implemented with two minor changes to the off-the-shelf RMSprop optimizer. Vprop also reduces the memory requirements of Black-Box Variational Inference by half. We derive Vprop using the conjugate-computation variational
   inference method, and establish its connections to Newton's method, natural-gradient methods, and extended Kalman filters.
 Overall, this paper presents Vprop as a principled, computationally-efficient, and easy-to-implement method for Bayesian deep learning.
 %By exploiting these new connections between optimization methods and approximate Bayesian inference methods, this paper proposes computationally-efficient methods for Bayesian deep learning that are also easy to implement. 
\end{abstract}

\section{Introduction}
Existing approaches for variational inference (VI), such as Black-Box Variational Inference (BBVI) \citep{ranganath2013black}, exploit stochastic-gradient methods to obtain simple and general implementations. Such approaches are widely applicable, but their implementation often requires significant changes to existing code-bases. For example, to implement BBVI for Bayesian neural networks, parameters have to be replaced with random variables, and the optimization objective is changed to the variational lower bound.
In this paper we propose a method for Gaussian variational approximations which simplifies the above by exploiting a connection between VI and modern optimization literature. 
We are able to implement variational inference by making two minor changes to the off-the-shelf RMSprop optimizer. A summary is given in Figure \ref{fig:summary}.

Our approach enables the VI implementation to lie entirely within the optimization procedure, and allows a plug-and-play of deterministic models such as neural networks.
By simply running the existing code-base of an optimizer, we surprisingly recover the optimum of the variational lower bound.
%The models' parameters are treated as random variables, and inference is performed on the Bayesian equivalent to the deterministic model.
%In the case of neural networks, the model's loss function is used as the negative log-likelihood in a KL objective in our optimizer, yielding a biased optimum compared to RMSprop. This biased optimum is the variational posterior mean, and the learnt step-sizes in our RMSprop variant form the variational posterior variance.
Apart from this marvellous connection between variational inference and modern optimization literature, this view also reduces the memory requirement of BBVI by half.
%(a closely-related connection was previously highlighted, for the case of gradient descent, in \citet{Opper:09})
This work provides a new software design paradigm to the field of Bayesian deep learning, and extends on recent ideas for efficient Bayesian approximations using deep-learning methodologies \citep{Gal2015Rapid, Gal2016Uncertainty, mandt2017stochastic}. For the latter part, we also establish connections of our method to Newton's method, natural-gradient methods, and extended Kalman filters.

\section{Optimization Algorithms for Variational Inference}
Continuous optimization algorithms are extremely popular in machine learning, e.g., given a supervised-learning problem with output vector $\vy$ and input matrix $\vX$, we can estimate model parameters $\vtheta$ by minimizing the negative log-likelihood: $f(\vtheta) := - \log p(\vy|\vX,\vtheta)$. This is the maximum-likelihood (ML) estimation, a popular method to fit complex models such as deep neural networks. This procedure scales well to large data and complex models, and the success of these models
is partly due to the existence of efficient implementations of optimization methods such as RMSprop \citep{hintonTieleman}, AdaGrad \citep{duchi2011adaptive}, and Adam \citep{kingma2014adam}.  

Variational inference (VI) methods hope to exploit optimization methods to approximate the posterior distribution $p(\vtheta|\vy,\vX)$, which often involves cumbersome integration. The problem of integration is fundamentally more difficult than finding a point estimate, and the key idea in VI is to convert the integration problem to an optimization problem. A common approach is to approximate the normalizing constant $p(\vy|\vX)$ of the posterior by finding an approximate distribution $q(\vtheta)$ that maximizes a
lower bound to it. For example, if we assume a Gaussian prior $p(\vtheta):=\gauss(\vtheta| 0, \vI/\lambda)$ with $\lambda>0$, we can approximate the posterior by a Gaussian distribution, $q(\vtheta) :=
\gauss(\vtheta|\vmu,\diag(\vsigma^2))$ with mean $\vmu$ and variance $\vsigma^2$. We do so by solving the following optimization problem:  
   \begin{align}
      \log p(\vy|\vX) &\ge \max_{\boldsymbol{\mu},\boldsymbol{\sigma}} \,\, \myexpect_q \sqr{ \log p(\vy|\vX,\vtheta) + \log p(\vtheta) -\log q(\vtheta)} \,\, := \mathcal{L}(\vmu,\vsigma^2). \label{eq:elbo}
	\end{align}
   Optimization algorithms can now be applied to solve this problem, and their efficiency can be exploited to perform approximate Bayesian inference. 

   Despite this reformulation, the implementation of an optimization algorithm for VI differs significantly from those used for ML estimation. For example, consider the Black-Box Variational Inference (BBVI) method \citep{ranganath2013black} which is one of the simplest approaches to optimize $\mathcal{L}$. BBVI employs the following simple stochastic-gradient update:
	\begin{align}
      \textrm{BBVI }: \quad \vmu_{t+1} = \vmu_t + \rho_t \sqr{\sg{\mathcal{L}_t}{\mu} }, 
      \quad\quad
	\vsigma_{t+1} &= \vsigma_t + \rho_t \sqr{ \sg{\mathcal{L}_t}{\sigma} } , \label{eq:vsgd_cov}
	\end{align}
	where $\rho_t>0$ is a step size at iteration $t$, $\widehat{\nabla}$ denotes an unbiased stochastic-gradient estimate, and $\nabla \mathcal{L}_t$ denotes the gradient at the value of the iterate at iteration $t$.
   These updates are simple and general, but their implementation differs significantly from adaptive-gradient methods (e.g., RMSprop; see Figure \ref{fig:rmsprop} for a pseudo-code).

A major challenge with BBVI is the large number of parameters which need to be optimized. Compared to the ML estimate, BBVI doubles the number of parameters (in the best case; when a full covariance matrix is used this becomes quadratic in the original problem size). This is because BBVI optimizes not only a single point estimate, but the parameters of an entire distribution. If an adaptive optimization scheme (like RMSprop) were to be used for BBVI to adapt step-sizes, the memory requirements
would have been doubled yet again in order to maintain a step-size for each of the distribution's parameters, i.e., two scaling vectors for $\vmu$ and $\vsigma$ respectively, instead of just one vector $\vs$ in standard RMSprop as shown in Figure \ref{fig:rmsprop}. This can become prohibitively expensive.
% The main differences arises from the fact that BBVI optimizes the parameters of a distribution instead of the parameter itself.
%    As a result of this is the number of parameters to be optimized are two times that of RMSprop.
%    Therefore, when an adaptive scheme is used to select step-sizes in BBVI, we need more memory to maintain separate scaling vectors for $\vmu$ and $\vsigma$.
   Further, adaptation is also tricky since $\vmu$ and $\vsigma$ are two fundamentally different quantities with different units and their step-sizes might require different types of tuning.
%For such updates, it is better to use a natural-gradient method (see \cite{hoffman2013stochastic} for a discussion on why natural-gradient is better suited to optimize the parameters of a distribution).
   A final difference is that BBVI requires gradients with respect to $\vsigma$ which typically demands a different implementation than the naive one to avoid numerical issues. 
%    To sum up, the implementation of methods such as BBVI, even though simple, would differ significantly from that of optimization algorithms such as RMSprop.

   In the next section we present perhaps surprising results, demonstrating a formulation of fully-factorized VI whose implementation is almost identical to that of the adaptive step-size RMSprop. Due to this similarity, we call our algorithm Vprop. As shown in Fig. \ref{fig:summary}, Vprop can be implemented with two minor changes to the implementation of RMSprop. We derive Vprop using a natural-gradient method of \cite{khan2017conjugate} called the conjugate-computation
   variational inference (CVI). Natural-gradient methods are preferable when optimizing parameters of a distribution
   \citep{hoffman2013stochastic} and our method has these desired theoretical properties as well.
   We establish additional connections to Newton's method, natural-gradient methods for continuous optimization, and extended Kalman filtering for approximate Bayesian inference.
   These connections make Vprop a principled approach for VI which is not only computationally-efficient but is also easy to implement.
   Empirical results on logistic regression and deep neural networks show that Vprop can indeed perform as well as BBVI but is much simpler to implement. 

	\begin{figure}[!t]
      \fbox{\subfigure[RMSprop update at $\vtheta=\vmu$ to find the maximum-likelihood estimate by minimizing $f(\vtheta):= -\log p(\vy|\vX,\vtheta)$]{
         \label{fig:rmsprop}
			\begin{minipage}{.35\textwidth}
				\begin{algorithmic}[1]
					\STATE $\vtheta \leftarrow \vmu$
					\STATE $\vg \leftarrow \nabla_{\theta} f(\vtheta)$
					\STATE $\vs \leftarrow (1-\beta) \vs + \beta (\vg.*\vg)$
					\STATE $\vmu \leftarrow \vmu - \alpha (\vg ./  \sqrt{\vs + \delta})$
				\end{algorithmic}
				\vspace{.1in}
			\end{minipage}
		}}
		\hfill
      \fbox{\subfigure[Vprop-1 update to find a Gaussian variational-distribution $q(\vtheta)$ that minimizes variational lower-bound (mean of $q$ is at $\vmu$ and its variance is equal to $1./(\vs+\lambda)$).]{
            \label{fig:vprop-1}
			\begin{minipage}{0.5\textwidth}
				\begin{algorithmic}[1]
					\STATE $\vtheta \leftarrow \vmu {\color{red} + \vepsilon./\sqrt{\vs + \lambda}}$ where $\vepsilon \sim \gauss(\vepsilon| 0,\vI)$
					\STATE $\vg \leftarrow \nabla_{\theta} f(\vtheta)$
					\STATE $\vs \leftarrow (1-\beta) \vs + \beta (\vg.* \vg)$ 
					\STATE $\vmu \leftarrow \vmu - \alpha ((\vg {\color{red} +  \lambda\vmu})./  {\color{red}(\vs +  \lambda)})$
				\end{algorithmic} 
				\vspace{.1in}
			\end{minipage}
		}}
      \caption{This figure compares the pseudo-code of RMSprop (left) and Vprop, our RMSprop variant used for variational inference (right). Differences between the two are highlighted in red. RMSprop optimizes the log-likelihood to find a maximum-likelihood estimate. For RMSprop, $\vmu$ is the current parameter vector, $\vs$ is the scaling vector, $\delta>0$ is a small constant, and $\alpha$ and $\beta$ are step-sizes. On the other hand, Vprop optimizes the variational lower bound \eqref{eq:elbo} to estimate a Gaussian variational-distribution with mean $\vmu$ and variance $\vsigma^2$.
         Here, $\lambda$ is the prior precision and the variance $\vsigma^2$ is obtained by setting $\vsigma^2 \leftarrow 1./(\vs+\lambda)$. The code of Vprop differs from RMSprop in lines 1 and 4 (highlighted in red). In line 1 in Vprop we add noise to the current parameter $\vmu$ which is equivalent to sampling $\vmu$ from the variational distribution. In line 4, a term $\lambda\vmu$ is added to the gradient, the scaling term is \emph{not} raised to the power $1/2$,
			and $\delta$ is set to be equal to $\lambda$. The code in the right is called Vprop-1 since it uses only one random sample. With just two lines of change in RMSprop, Vprop performs variational inference.}
%     \vspace{2mm}
		\label{fig:summary}
	\end{figure}
    
	\section{From Variational Inference to an RMSprop Variant}
   We will derive Vprop using the conjugate-computation variational inference (CVI) method proposed by \cite{khan2017conjugate}. CVI is a natural-gradient method for VI and when applied to \eqref{eq:elbo}, results in the following update (see Appendix \ref{app:derivation_van} for a proof): 
	\begin{align}
	\textrm{CVI: } \quad \vmu_{t+1} &= \vmu_{t} + \beta_t \,\, \vsigma_{t+1}^2 \circ \sqr{\nabla_\mu \mathcal{L}}, 
      \quad
	\vsigma_{t+1}^{-2} = \vsigma_t^{-2} - \,\, 2\beta_t \,\, \sqr{ \nabla_{\sigma^2} \mathcal{L}},  \label{eq:Van_prec_0}
	\end{align}
   where $\circ$ denotes element-wise multiplication of two vectors.
   These updates are natural-gradient updates and differ from BBVI update of \eqref{eq:vsgd_cov} in two main aspects. First, these updates use gradients with respect to the variance $\vsigma^2$ to update the precision $\vsigma^{-2}$, while BBVI uses the gradient w.r.t.~ $\vsigma$ to update the standard-deviation $\vsigma$.
   Second, the update for $\vmu$ is an adaptive update because the step-size $\beta_t$ is scaled by the variance. 
   As we show next, these two differences enable the implementation of CVI using an RMSprop variant, which is not possible for BBVI.  

   Vprop can be derived from CVI in two steps. First, we use Bonnet's and Price's theorem \citep{rezende2014stochastic} to express the gradients with respect to $\vmu$ and $\vsigma^2$ in terms of gradient and Hessian of $f$ respectively.
Specifically, we use the following two identities \citep{Opper:09}:
   \begin{align}
      \nabla_{\mu} \myexpect_q \sqr{f(\vtheta) } = \myexpect_q \sqr{ \nabla_\theta f(\vtheta)}, \quad\quad \nabla_{\sigma^2} \myexpect_q \sqr{f(\vtheta) } = \half \myexpect_q \sqr{ \diag( \nabla^2_{\theta\theta} f(\vtheta) ) } \label{eq:bonnet}
   \end{align}
   where $\diag(\vA)$ extracts the diagonal of $\vA$.
   Using these, we can rewrite the gradients as follows:
	\begin{align}
      \nabla_\mu \mathcal{L} &= \nabla_\mu \myexpect_{q} \sqr{ -f(\vtheta) + \log p(\vtheta) - \log q(\vtheta)} = -\myexpect_{q}\sqr{ \nabla_\theta f(\vtheta)} - \lambda\vmu,  \label{eq:mfgradmu}\\
\nabla_{\sigma^2} \mathcal{L} &= -\half \myexpect_{q} \crl{\diag\sqr{ \nabla^2_{\theta\theta} f(\vtheta) }} - \half \lambda\vone + \half\vsigma^{-2} ,\label{eq:mfgradsig}
	\end{align}
   where $\vone$ is a vector of ones and $\vsigma^{-2}$ denotes element-wise inverse square of the elements of the vector $\vsigma$.
   Using the above, we can rewrite the CVI updates as the following:
	%\begin{align}
	%\textrm{VON: } \quad \vmu_{t+1} &= \vmu_{t} - \beta_t \,\, \vP_{t+1}^{-1} \myexpect_{q_t} \sqr{ \nabla_\theta f(\vtheta)}   \label{eq:Von_mean_0},\\
	%\vP_{t+1} &= (1-\beta_t) \vP_t +  \beta_t \,\, \myexpect_{q_t} \sqr{ \nabla^2_{\theta\theta} f(\vtheta) }.  \label{eq:Von_prec_0}
	%\end{align}
   \begin{align}
	\vmu_{t+1} &= \vmu_{t} - \beta_t \,\, \vsigma_{t+1}^2 \circ \crl{ \myexpect_{q_t} \sqr{ \nabla_\theta f(\vtheta)} + \lambda\vmu_t },   \label{eq:Vond_mean_0}\\
      \vsigma_{t+1}^{-2} &= (1-\beta_t) \vsigma_t^{-2} +  \beta_t \,\, \crl{ \myexpect_{q_t} \sqr{ \diag\rnd{ \nabla^2_{\theta\theta} f(\vtheta) } } + \lambda\vone},  \label{eq:Vond_prec_0}
	\end{align}
   where $q_t := \gauss(\vtheta|\vmu_t, \vsigma_t^2)$ is the variational distribution at iteration $t$.
   %The above updates maintain an \emph{online} estimate of the precision vector $\vsigma_t^2$ by using a moving-average of the past \emph{averaged Hessians}, i.e., Hessians averaged over $\vtheta \sim q_t$. 
   %Also, these updates are similar to Newton's method because the search-direction for the mean is obtained by scaling the \emph{averaged gradients}, i.e., gradients averaged over $\vtheta\sim q_t$, by $\vsigma_t^2$ which contains the second-order information.
   %Due to these connections, we call this method the \emph{Variational Online-Newton with Diagonal-approximation method} or VON-D method. 
   %CVI updates are very similar to the online natural-gradient updates discussed in \cite{YannOllivier1703.00209} where a Gauss-Newton approximation of the Hessian $\nabla_{\theta\theta}^2 f(\vtheta)$ is employed.

   These updates require computation of a Hessian which might be computationally difficult.
   The second step in Vprop derivation is to replace the Hessian by a Gauss-Newton approximation \citep{bertsekas1999nonlinear}. This approximation is also numerically useful when the Hessian is not positive semi-definite, which is typically the case when $f$ is parameterized by a neural network. Using the Gauss-Newton approximation, we can simplify the precision update to the following: 
	\begin{align}
      \vsigma_{t+1}^{-2} &= (1-\beta_t) \vsigma_t^{-2} +  \beta_t \,\, \sqr{ \myexpect_{q_t} [ \rnd{\nabla_{\theta} f(\vtheta) }^2 ] + \lambda\vone}.
	\end{align}
   By defining ${\vs}_t := \vsigma_t^{-2} - \lambda\vone$, we can rewrite the update as follows, which we call Vprop:
	%We can rewrite the updates in terms of ${\vs}_t = \vp_t - \lambda\vone$, as shown below:
	%\begin{align}
%	\textrm{Vprop: } \quad \vmu_{t+1} &= \vmu_{t} - \beta_t \,\,({\vs}_{t+1} + \lambda\vone)^{-1} \circ  \crl{ \myexpect_{q_t} \sqr{ \nabla_\theta f(\vtheta) } + \lambda \vmu_t } ,\\
%	{\vs}_{t+1} &= (1-\beta_t) {\vs}_t +  \beta_t \,\, \myexpect_{q_t} \sqr{ \rnd{ \nabla_\theta f(\vtheta) }^2  }.% 
	%\end{align}
%
   %By defining ${\vs}_t := \vsigma_t^{-2} - \lambda\vone$ and after a few rearrangements, we arrive at the following update:
   \begin{align}
	\textrm{Vprop: } \quad \vmu_{t+1} &= \vmu_{t} - \beta_t \,\,({\vs}_{t+1} + \lambda\vone)^{-1} \circ  \crl{ \myexpect_{q_t} \sqr{ \nabla_\theta f(\vtheta) } + \lambda \vmu_t } , \label{eq:Vprop_mean}\\
	{\vs}_{t+1} &= (1-\beta_t) {\vs}_t +  \beta_t \,\, \myexpect_{q_t} \sqr{ \rnd{ \nabla_\theta f(\vtheta) }^2  }.  \label{eq:Vprop_prec}
	\end{align}
   %A detailed derivation is given in Appendix \ref{app:vprop_deriv}.
   Approximating the expectation using one sample $\vtheta_t \sim \gauss(\vtheta|\vmu_t,1/(\vs_t+\lambda\vone))$, we get the variant of Vprop we call Vprop-1 (also shown in Fig. \ref{fig:summary}):
   \begin{align}
	\textrm{Vprop-1: } \quad \vmu_{t+1} &= \vmu_{t} - \beta_t \,\,({\vs}_{t+1} + \lambda\vone)^{-1} \circ  \sqr{  \nabla_\theta f(\vtheta_t) + \lambda \vmu_t } , \label{eq:Vprop_mean}\\
	{\vs}_{t+1} &= (1-\beta_t) {\vs}_t +  \beta_t \,\, \sqr{ \nabla_\theta f(\vtheta_t) }^2  .  \label{eq:Vprop_prec}
	\end{align}
   We can see the similarity to RMSprop by comparing its update to Vprop-1 as shown in Fig. \ref{fig:summary}.
	Both algorithms use a running sum of the square of the gradient to compute the scaling vector $\vs_t$.
	The algorithms differ in only two lines of code with three major differences.
	First, Vprop uses samples from $q$ to compute the gradient. This enables a local-exploration around the mean which is useful for uncertainty computation.
	Second, RMSprop raises the scaling vector to a power of $1/2$, while Vprop does not.
	Third, Vprop adds the term $\lambda\vmu_t$ to the gradient in the mean update.
	These three differences result in the surprising conversion of RMSprop, which maximizes the log-likelihood, to Vprop, which maximizes the variational lower bound.

   We can also derive a deterministic version of Vprop called Vprop-0 which is a bit more similar to RMSprop than Vprop-1. In Vprop-0, instead of using MC samples, we approximate the expectation using a first-order delta approximation $\myexpect_q  \sqr{ \vg(\vtheta)} \approx \vg(\vmu)$ and $\myexpect_q  \sqr{( \vg(\vtheta)^2 } \approx \vg(\vmu)^2$.
   This gives us the following update:
	\begin{align}
      \textrm{Vprop-0: } \quad \vmu_{t+1} &= \vmu_{t} - \beta_t \,\,({\vs}_{t+1} + \lambda\vone)^{-1} \circ \sqr{ \nabla_\theta f(\vmu_t) + \lambda\vmu_t} , \label{eq:vprop0-mean} \\
      {\vs}_{t+1} &= (1-\beta_t) {\vs}_t +  \beta_t \,\, \sqr{\nabla_\theta f(\vmu_t)}^2 \label{eq:vprop0-prec}
	\end{align}
    with $\vg(\vmu_t)$ the gradient of $f$ at $\vmu_t$. Our empirical results show that Vprop-0 performs worse than Vprop-1, establishing the importance of local exploration obtained by using samples from $q$.

	%We call these updates Vprop-0 because they do not use any samples to estimate the expectation. 

    \section{Connections to Newton's Method and Natural-Gradient Methods}
    In this section, we consider extensions to non mean-field variational distribution, i.e., when the covariance $\vSigma$ of $q(\vtheta)$ is not a diagonal matrix but a full matrix. For this case, we show that our algorithm is a second-order method and is related to an online version of Newton's method. By making a Gauss-Newton approximation to the Hessian, we establish connections to online natural-gradient method and extended Kalman filtering (EKF) method described in
    \cite{YannOllivier1703.00209}. The results presented in this section connect methods from three fields: variational inference, continuous optimization, and approximate Bayesian filtering.

The results derived in this section are similar to another work by \cite{2017arXiv171105560E} who derive Newton-type methods for general purpose optimization. Our derivations are similar to theirs, but our results are about variational inference which is a different problem than the one considered in \cite{2017arXiv171105560E}.

        The CVI algorithm for the fully-correlated case can be derived in a similar way to the derivation given in Appendix \ref{app:derivation_van}. The resulting updates are shown below:
   \begin{align}
	\textrm{CVI-Full: } \quad \vmu_{t+1} &= \vmu_{t} + \beta_t \,\, \vSigma_{t+1} \sqr{\nabla_\mu \mathcal{L}}, 
      \quad
      \vSigma_{t+1}^{-1} = \vSigma_t^{-1} - \,\, 2\beta_t \,\, \sqr{ \nabla_{\Sigma} \mathcal{L}},  \label{eq:Van_prec_0}
	\end{align}
   Using the identities given in \eqref{eq:bonnet}, gradient expressions similar to \eqref{eq:mfgradmu} and \eqref{eq:mfgradsig}, and one MC sample approximation, we can rewrite the updates in terms of the gradient and Hessian of $f(\vtheta)$ as shown below:
    \begin{align}
       \textrm{VON-1: } \quad \vmu_{t+1} &= \vmu_{t} - \beta_t \,\, \rnd{\vS_{t+1} + \lambda\vI}^{-1} \rnd{ \nabla_\theta f(\vtheta_t) + \lambda\vmu_t}   \label{eq:Von_mean_0},\\
       \vS_{t+1} &= (1-\beta_t) \vS_t +  \beta_t \,\,  \nabla^2_{\theta\theta} f(\vtheta_t) ,  \label{eq:Von_prec_0}
\end{align}
where $\vS_t = \vSigma_t^{-1} -\lambda\vI$ is the scaling matrix and $\vtheta_t$ is a sample from $q_t:= \gauss(\vtheta|\vmu_t, (\vS_t + \lambda\vI)^{-1})$. We refer to this update as the Variational Online-Newton (VON) method because it resembles an online version of Newton's method where the scaling matrix is estimated online (the number 1 indicates that expectations are approximated with one MC sample).
We can clearly see this resemblance by comparing VON to the update for Newton's method:
\begin{align}
   \textrm{Newton's method: } \vtheta_{t+1} &= \vtheta_t - \rho_t \sqr{\nabla^2_{\theta\theta} f(\vtheta_t)}^{-1} \sqr{\nabla_\theta f(\vtheta_t)}.
\end{align}
In VON, the scaling matrix is a moving-average of the past Hessians and each Hessian is evaluated at a sample from $q(\vtheta)$. The scaling matrix maintains an online estimate of the past curvature information, making the update an online second-order method.

A major difference in VON is that the gradients and Hessians are evaluated at the samples from $q(\vtheta)$ instead of the current parameter.
This enables a local exploration around the current parameter values which is expected to improve the performance by avoiding some local minima (see an example of such local-minima avoidance in \cite{2017arXiv171105560E}). This difference shows the potential benefits obtained when using an approximate Bayesian method instead of a point-estimate method.

Like the Vprop derivation discussed in the previous section, if we use a Gauss-Newton approximation for the Hessian, the resulting updates are similar to the online natural-gradient descent:
\begin{align}
   \textrm{VONG-1: } \quad \vmu_{t+1} &= \vmu_{t} - \beta_t \,\, \rnd{\vS_{t+1} + \lambda\vI}^{-1} \rnd{ \nabla_\theta f(\vtheta_t) + \lambda\vmu_t}   \label{eq:Vog_mean_0},\\
   \vS_{t+1} &= (1-\beta_t) \vS_t +  \beta_t \,\,  \nabla_{\theta} f(\vtheta_t)  \sqr{\nabla_{\theta} f(\vtheta_t)}^T.  \label{eq:Vog_prec_0}
\end{align}
Due to this similarity, we call it the Variational Online Natural-Gradient (VONG) algorithm.

The VONG update is very similar to the regularized online natural-gradient step discussed in Proposition 4~in~\cite{YannOllivier1703.00209}. There, the Gaussian prior over $\vtheta$ is used as a Bayesian regularizer of the Fisher matrix. In VONG, the regularization naturally arises as a result of performing approximate inference in a Bayesian model.
\cite{YannOllivier1703.00209} also show that their online natural-gradient algorithm is equivalent to extended Kalman filters (EKF).
Therefore, VONG is also closely related to EKF.

VONG differs from the method of \cite{YannOllivier1703.00209} in that VONG uses an empirical estimate of the Fisher matrix obtained by using the observed data $\vy$. The method of \cite{YannOllivier1703.00209}, on the other hand, approximates the Fisher matrix by using an average over $p(y|\vx,\vtheta)$. Traditionally, natural-gradient methods have relied on empirical estimates \citep{amari1998natural}, but
recent studies, such as \cite{pascanu2013revisiting}, have shown that using averaging leads to an unbiased estimate and performs better. 
Vprop updates are amenable to such modifications, although it is not clear if this is a valid step to perform variational inference.

The above connections of our new VI methods to optimization algorithms are useful in establishing general connections between the two distinct fields. VI methods optimize in the space of variational distribution $q$ and are fundamentally different from continuous optimization methods which optimize in the space of parameter $\vtheta$. Yet, as we show in this paper, it is possible to design VI methods by only slightly modifying existing optimization methods. Some other recent works have shown similar types of results \citep{Gal2015Rapid, Gal2016Uncertainty, mandt2017stochastic}. Such results are very encouraging and further promote the use of optimization methods as a tool to design scalable algorithms for Bayesian deep learning.

\section{Experimental Results}
\label{app:results}

\begin{figure}
	\centering
   \subfigure{\includegraphics[width=2.7in,height = 1.25in]{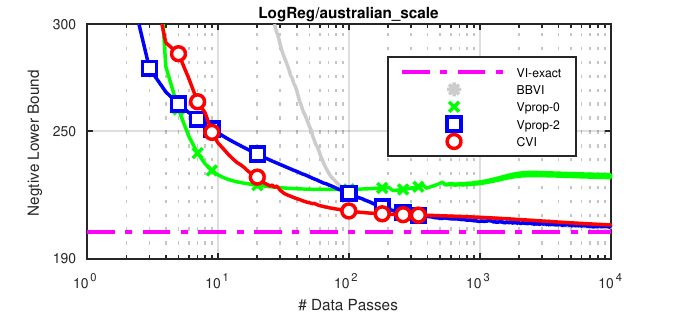}}
   \subfigure{\includegraphics[width=2.7in,height = 1.25in]{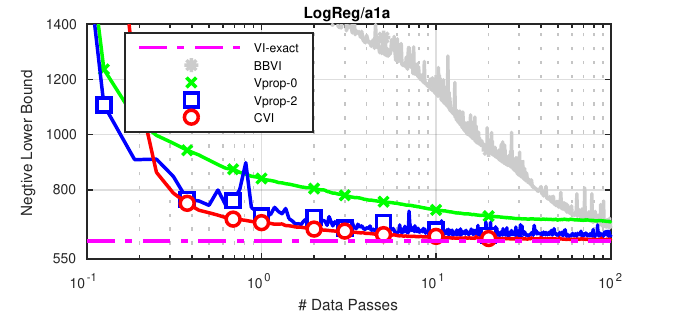}}
   \subfigure{\includegraphics[width=2.7in,height = 1.25in]{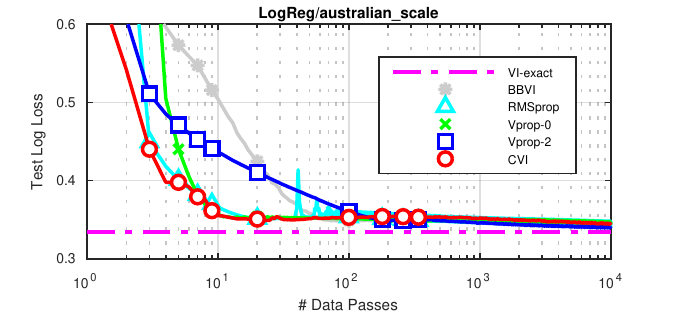}}
   \subfigure{\includegraphics[width=2.7in,height = 1.25in]{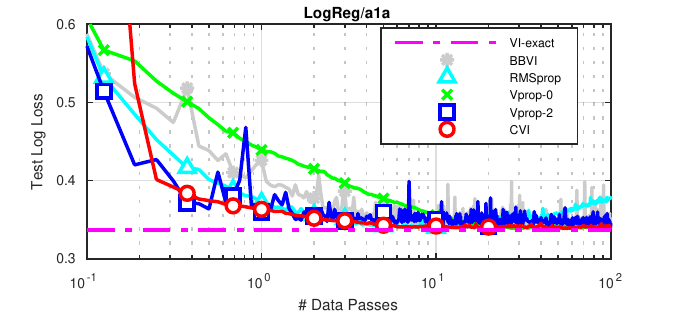}}
   \caption{Results on logistic regression. Left column shows results for the Australian-Scale dataset ($N= 345, D=15$, $\lambda=10^{-5}$) where we plot ELBO on training data and log-loss on test data, respectively, versus number of data passes. Right column shows the same for the `a1a' dataset ($N=1605, D=123, \lambda = 2.8072$). `VI-exact' is the ground truth obtained by using LBFGS. BBVI is the update \eqref{eq:vsgd_cov} with constant step-sizes. We also compare to `CVI' using update
      \eqref{eq:Vond_mean_0}-\eqref{eq:Vond_prec_0} with 10 MC samples and exact Hessian computation. For our method, we use `Vprop-2' using update
      \eqref{eq:Vprop_mean}-\eqref{eq:Vprop_prec} with 2 MC samples, and `Vprop-0' implementing update \eqref{eq:vprop0-mean}-\eqref{eq:vprop0-prec}. We see that they all converge either faster than BBVI or at the same rate, while enabling much simpler implementation. CVI uses exact Hessian computation, while Vprop-2 does not and still performs very similar. We also show the log-loss performance of the pure RMSprop method outlined in Fig. \ref{fig:rmsprop}. This method does not optimize ELBO
   or compute uncertainty, but performs well.}
   \label{fig:logreg}
\end{figure}

\begin{figure}
   \subfigure{\includegraphics[width=2.7in,height = 1.25in]{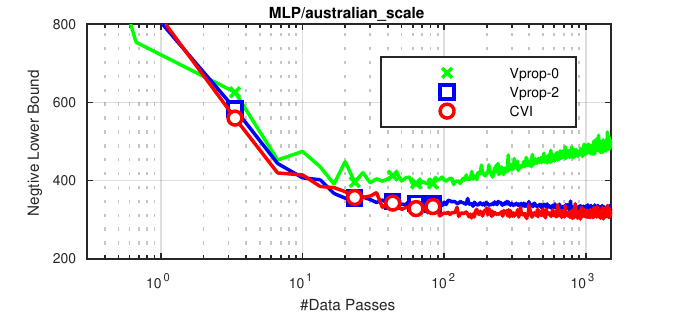}}
   \subfigure{\includegraphics[width=2.7in,height = 1.25in]{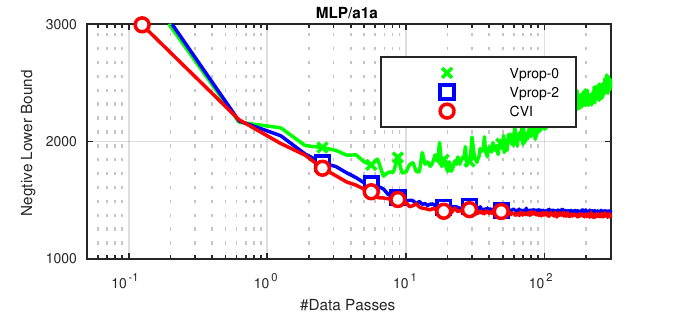}}
   \subfigure{\includegraphics[width=2.7in,height = 1.25in]{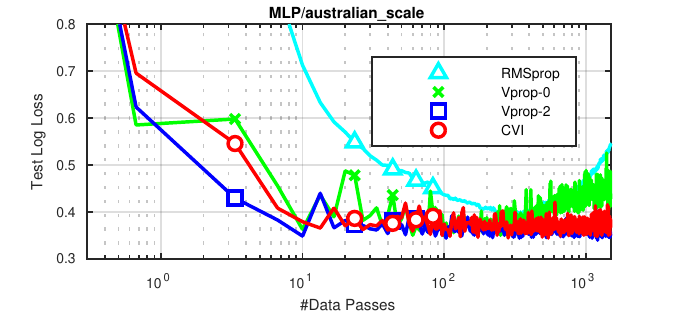}}
   \subfigure{\includegraphics[width=2.7in,height = 1.25in]{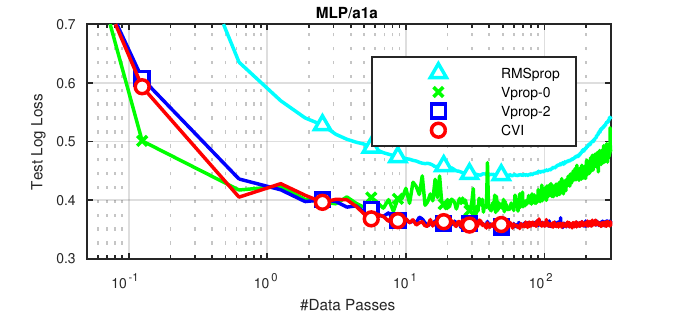}}
   \caption{Same comparison as Figure \ref{fig:logreg} but by using Multi-Layer Perceptron (MLP) with two hidden layers, 10 units each ($\lambda = 1$). We do not have the ground truth for MLP since computing exact ELBO is difficult. For MLP, Vprop-2 performs almost the same as CVI which is expected since, for nonconvex objectives, Gauss-Newton is usually a good numerical approximation. Both Vprop-0 and RMSprop start overfitting after some iterations, but methods that use MC sampling,
      i.e., Vprop-2 and CVI, do not. We
      conjecture that this is because the MC sampling gives unbiased gradients that optimize ELBO, while the other two methods do not do so and overfit.}
      \label{fig:mlp}
\end{figure}

In this section, we present results to establish that Vprop gives comparable performance to existing VI methods.
   We show results logistic regression and multi-layer perceptron in Figure \ref{fig:logreg} and \ref{fig:mlp} respectively. Our results show that Vprop performs as well as CVI despite using the Gauss-Newton approximation. We also show that RMSprop overfits on these datasets, perhaps due to small data-size. Vprop-0 also performs badly but slightly better than RMSprop. We believe that the worse performance is because Vprop-0 does not use samples from $q$ which might lead to overfitting.
   The slightly better performance of Vprop-0 compared to RMSprop is perhaps because it does not use the square root of the scaling vector.


\section{Discussion and Future Works}
We proposed Vprop, a Gaussian VI method that can be implemented with two minor changes to the off-the-shelf RMSprop optimizer. The memory requirement of Vprop is half of that required by BBVI. In addition, Vprop is an approximate natural-gradient VI method and inherits many good theoretical properties of natural-gradient methods. 
We show that Vprop is related to online versions of Newton's method and the natural-gradient method, and also to extended Kalman filters. Vprop is a principled and computationally-efficient approach for VI, and is also an easy-to-implement method.

We have provided experimental evidence on small models and datasets. In our experiments, Vprop beats BBVI with other adaptive-gradient methods (not presented in the plots), but further experiments are required to confirm this. In the future, we plan to do extensive comparisons on larger problems and compare Vprop to many other existing methods. We also hope to compare to a version of Vprop with momentum and to RMSprop with momentum. We also plan to compare Vprop to other existing methods such as Bayesian Dropout and other black-box methods for VI.

{\bf Acknowledgement:} We thank Wu Lin (RIKEN) and Didrik Nielsen (RIKEN) for useful discussions. We also thank anonymous reviewers for their useful feedback.

\bibliography{paper}

\begin{thebibliography}{16}
\providecommand{\natexlab}[1]{#1}
\providecommand{\url}[1]{\texttt{#1}}
\expandafter\ifx\csname urlstyle\endcsname\relax
  \providecommand{\doi}[1]{doi: #1}\else
  \providecommand{\doi}{doi: \begingroup \urlstyle{rm}\Url}\fi

\bibitem[Amari(1998)]{amari1998natural}
Shun-Ichi Amari.
\newblock Natural gradient works efficiently in learning.
\newblock \emph{Neural computation}, 10\penalty0 (2):\penalty0 251--276, 1998.

\bibitem[Bertsekas(1999)]{bertsekas1999nonlinear}
Dimitri~P Bertsekas.
\newblock \emph{Nonlinear programming}.
\newblock Athena Scientific, 1999.

\bibitem[Duchi et~al.(2011)Duchi, Hazan, and Singer]{duchi2011adaptive}
John Duchi, Elad Hazan, and Yoram Singer.
\newblock Adaptive subgradient methods for online learning and stochastic
  optimization.
\newblock \emph{The Journal of Machine Learning Research}, 12:\penalty0
  2121--2159, 2011.

\bibitem[Gal(2015)]{Gal2015Rapid}
Yarin Gal.
\newblock Rapid prototyping of probabilistic models: Emerging challenges in
  variational inference.
\newblock In \emph{Advances in Approximate Bayesian Inference workshop, NIPS},
  2015.

\bibitem[Gal(2016)]{Gal2016Uncertainty}
Yarin Gal.
\newblock \emph{Uncertainty in Deep Learning}.
\newblock PhD thesis, University of Cambridge, 2016.

\bibitem[Hoffman et~al.(2013)Hoffman, Blei, Wang, and
  Paisley]{hoffman2013stochastic}
Matthew~D Hoffman, David~M Blei, Chong Wang, and John Paisley.
\newblock Stochastic variational inference.
\newblock \emph{The Journal of Machine Learning Research}, 14\penalty0
  (1):\penalty0 1303--1347, 2013.

\bibitem[Khan \& Lin(2017)Khan and Lin]{khan2017conjugate}
Mohammad~Emtiyaz Khan and Wu~Lin.
\newblock Conjugate-computation variational inference: Converting variational
  inference in non-conjugate models to inferences in conjugate models.
\newblock \emph{arXiv preprint arXiv:1703.04265}, 2017.

\bibitem[{Khan} et~al.(2017){Khan}, {Lin}, {Tangkaratt}, {Liu}, and
  {Nielsen}]{2017arXiv171105560E}
Mohammad~Emtiyaz {Khan}, Wu~{Lin}, Voot {Tangkaratt}, Zuozhu {Liu}, and Didrik
  {Nielsen}.
\newblock {{Variational Adaptive-Newton Method for Explorative Learning}}.
\newblock \emph{ArXiv e-prints}, November 2017.

\bibitem[Kingma \& Ba(2014)Kingma and Ba]{kingma2014adam}
Diederik Kingma and Jimmy Ba.
\newblock {Adam: A method for stochastic optimization}.
\newblock \emph{arXiv preprint arXiv:1412.6980}, 2014.

\bibitem[Mandt et~al.(2017)Mandt, Hoffman, and Blei]{mandt2017stochastic}
Stephan Mandt, Matthew~D Hoffman, and David~M Blei.
\newblock Stochastic gradient descent as approximate bayesian inference.
\newblock \emph{arXiv preprint arXiv:1704.04289}, 2017.

\bibitem[Ollivier(2017)]{YannOllivier1703.00209}
Yann Ollivier.
\newblock Online natural gradient as a kalman filter, 2017.

\bibitem[Opper \& Archambeau(2009)Opper and Archambeau]{Opper:09}
M.~Opper and C.~Archambeau.
\newblock {The Variational {G}aussian Approximation Revisited}.
\newblock \emph{Neural Computation}, 21\penalty0 (3):\penalty0 786--792, 2009.

\bibitem[Pascanu \& Bengio(2013)Pascanu and Bengio]{pascanu2013revisiting}
Razvan Pascanu and Yoshua Bengio.
\newblock Revisiting natural gradient for deep networks.
\newblock \emph{arXiv preprint arXiv:1301.3584}, 2013.

\bibitem[Ranganath et~al.(2014)Ranganath, Gerrish, and
  Blei]{ranganath2013black}
Rajesh Ranganath, Sean Gerrish, and David~M Blei.
\newblock Black box variational inference.
\newblock In \emph{International conference on Artificial Intelligence and
  Statistics}, pp.\  814--822, 2014.

\bibitem[Rezende et~al.(2014)Rezende, Mohamed, and
  Wierstra]{rezende2014stochastic}
Danilo~Jimenez Rezende, Shakir Mohamed, and Daan Wierstra.
\newblock Stochastic backpropagation and approximate inference in deep
  generative models.
\newblock \emph{arXiv preprint arXiv:1401.4082}, 2014.

\bibitem[Tieleman \& Hinton(2012)Tieleman and Hinton]{hintonTieleman}
Tijmen Tieleman and Geoffrey Hinton.
\newblock {Lecture 6.5-{R}MSprop: Divide the gradient by a running average of
  its recent magnitude.}
\newblock \emph{COURSERA: Neural Networks for Machine Learning 4}, 2012.

\end{thebibliography}
\bibliographystyle{iclr2018_conference}

\appendix
\section{Derivation of CVI for Gaussian variational distribution}
\label{app:derivation_van}

Denote the mean parameters of $q_t(\vtheta)$ by $\vm_t := \{\vmu, \vmu^2 + \vsigma_t^{2}\}$. The mean parameter is equal to the expected value of the sufficient statistics $\vphi(\vtheta) := \{\vtheta, \vtheta^2\}$, i.e., $\vm_t := \myexpect_{q_t} [\vphi(\vtheta)] $. The mirror descent update at iteration $t$ is given by the solution to
\begin{align}
\vm_{t+1} 
&= \argmax_{\boldsymbol{m}} \left\langle \vm, \widehat{\nabla}_{m} \mathcal{L}_t \right\rangle - \frac{1}{\beta_t}\dkls{}{q}{q_t} \\
&= \argmin_{\boldsymbol{m}} \left\langle \vm, -\widehat{\nabla}_{m} \mathcal{L}_t \right\rangle + \frac{1}{\beta_t}\dkls{}{q}{q_t} \\
&= \argmin_{\boldsymbol{m}} \myexpect_q \left[ \left\langle \vphi(\vtheta), -\widehat{\nabla}_{m} \mathcal{L}_t \right\rangle + \log\left((q/q_t)^{1/\beta_t}\right) \right] \\
&= \argmin_{\boldsymbol{m}} \myexpect_q \left[ \log \frac{ \exp \left\langle \vphi(\vtheta), -\widehat{\nabla}_{m} \mathcal{L}_t \right\rangle q^{1/\beta_t} }{ q_t^{1/\beta_t}} \right] \\
&= \argmin_{\boldsymbol{m}} \myexpect_q \left[ \log \left( \frac{q^{1/\beta_t}}{q_t^{1/\beta_t} \exp \left\langle \vphi(\vtheta), \widehat{\nabla}_{m} \mathcal{L}_t \right\rangle } \right) \right] \\
&= \argmin_{\boldsymbol{m}} \frac{1}{\beta_t} \,\, \myexpect_q \left[ \log \left( \frac{q}{q_t \exp \left\langle \vphi(\vtheta), \beta_t \widehat{\nabla}_{m} \mathcal{L}_t \right\rangle } \right) \right] \\
&= \argmin_{\boldsymbol{m}}  \frac{1}{\beta_t} \,\,\mathbb{D}_{KL}\sqr{ q \| q_t \exp \left\langle \vphi(\vtheta), \beta_t \widehat{\nabla}_{m} \mathcal{L}_t \right\rangle /\mathcal{Z}_t} .
\end{align}
where $\mathcal{Z}$ is the normalizing constant of the distribution in the denominator which is a function of the gradient and step size.

Minimizing this KL divergence gives the update
\begin{equation}
q_{t+1}(\vtheta) \propto q_t(\vtheta)\exp \left\langle \vphi(\vtheta), \beta_t \widehat{\nabla}_{m} \mathcal{L}_t \right\rangle.
\end{equation}
By rewriting this, we see that we get an update in the natural parameters $\vlambda_t$ of $q_t(\vtheta)$, i.e.
\begin{equation}
\vlambda_{t+1} = \vlambda_t + \beta_t \widehat{\nabla}_{m} \mathcal{L}_t.
\end{equation}
Recalling that the mean parameters of a Gaussian $q(\vtheta) = \gauss(\vtheta|\vmu,\diag(\vsigma^2) )$ are $\vm^{(1)} = \vmu$ and $\vm^{(2)} = \vsigma^2 + \vmu^2$ and using the chain rule, we can express the gradient $\widehat{\nabla}_{m} \mathcal{L}_t$ in terms of $\vmu$ and $\vsigma^2$,
\begin{align}
    \widehat{\nabla}_{m^{(1)}} \mathcal{L} &= \widehat{\nabla}_{\mu} \mathcal{L} - 2\left[\widehat{\nabla}_{\sigma^2} \mathcal{L} \right] \circ \vmu \\
    \widehat{\nabla}_{m^{(2)}} \mathcal{L} &= \widehat{\nabla}_{\sigma^2} \mathcal{L}.
\end{align}
Finally, recalling that the natural parameters of a Gaussian $q(\vtheta) = \gauss(\vtheta|\vmu,\diag(\vsigma^2))$ are $\vlambda^{(1)} = \vsigma^{-2} \circ\vmu$ and $\vlambda^{(2)} = - \frac{1}{2}\vsigma^{-2}$, we can rewrite the CVI updates in terms of $\vmu$ and $\vsigma^2$,
\begin{align}
    \vsigma_{t+1}^{-2} &= \vsigma_t^{-2} - 2\beta_t \left[ \widehat{\nabla}_{\sigma^2} \mathcal{L}_t \right] , \\
    \vmu_{t+1} &= \vsigma_{t+1}^2\circ  \left[ \vsigma_t^{-2}\circ \vmu_t + \beta_t \left( \widehat{\nabla}_{\mu} \mathcal{L}_t - 2\left[\widehat{\nabla}_{\sigma^2} \mathcal{L}_t \right] \circ \vmu_t \right) \right] ,\\
    &= \vsigma_{t+1}^2 \circ \left( \vsigma_t^{-2} - 2\beta_t \left[ \widehat{\nabla}_{\sigma^2} \mathcal{L}_t \right] \right)\vmu_t + \beta_t\vsigma_{t+1}^2 \circ \left[ \widehat{\nabla}_{\mu} \mathcal{L}_t \right] , \\
    &= \vmu_t + \beta_t\vsigma_{t+1}^2 \circ \left[ \widehat{\nabla}_{\mu} \mathcal{L}_t \right].
\end{align}

\end{document}